# Image inpainting: A review

Omar Elharrouss[a], Noor Almaadeed[a], Somaya Al-Maadeed[a], Younes Akbari[a]

[a]*Department of Computer Science and Engineering, Qatar university, Doha, Qatar*

**Abstract**

Although image inpainting, or the art of repairing the old and deteriorated images, has been around for many years, it has gained even more popularity because of the recent development in image processing techniques. With the improvement of image processing tools and the flexibility of digital image editing, automatic image inpainting has found important applications in computer vision and has also become an important and challenging topic of research in image processing. This paper is a brief review of the existing image inpainting approaches we first present a global vision on the existing methods for image inpainting. We attempt to collect most of the existing approaches and classify them into three categories, namely, sequential-based, CNN-based and GAN-based methods. In addition, for each category, a list of methods for the different types of distortion on the images is presented. Furthermore, collect a list of the available datasets and discuss these in our paper. This is a contribution for digital image inpainting researchers trying to look for the available datasets because there is a lack of datasets available for image inpainting. As the final step in this overview, we present the results of real evaluations of the three categories of image inpainting methods performed on the datasets used, for the different types of image distortion. In the end, we also present the evaluations metrics and discuss the performance of these methods in terms of these metrics. This overview can be used as a reference for image inpainting researchers, and it can also facilitate the comparison of the methods as well as the datasets used. The main contribution of this paper is the presentation of the three categories of image inpainting methods along with a list of available datasets that the researchers can use to evaluate their proposed methodology against.

*Keywords:* image inpainting, CNN, GAN

## 1. Introduction

Nowadays, image is one of the most common forms of information that is used in every domain of life. In addition, it is a crucial tool for monitoring the security of people and objects. But the editing applications that can edit an image without leaving any traces,

*Email addresses:* `elharrouss.omar@gmail.com` (Omar Elharrouss), `n.alali@qu.edu.qa` (Noor Almaadeed), `s_alali@qu.edu.qa` (Somaya Al-Maadeed), `akbari_younes@semnan.ac.ir`(Younes Akbari)

*September 13, 2019*

pose a problem to the public trust and confidence. So, the need for an automatic system to detect and extract the real image present in the available is an urgent demand. Meanwhile, the availability of original image from the given image is heavily dependent on the extraction mechanism of the original image, hence object removal from images is one of the big concerns of the research and a hot topic for information security [1,2].

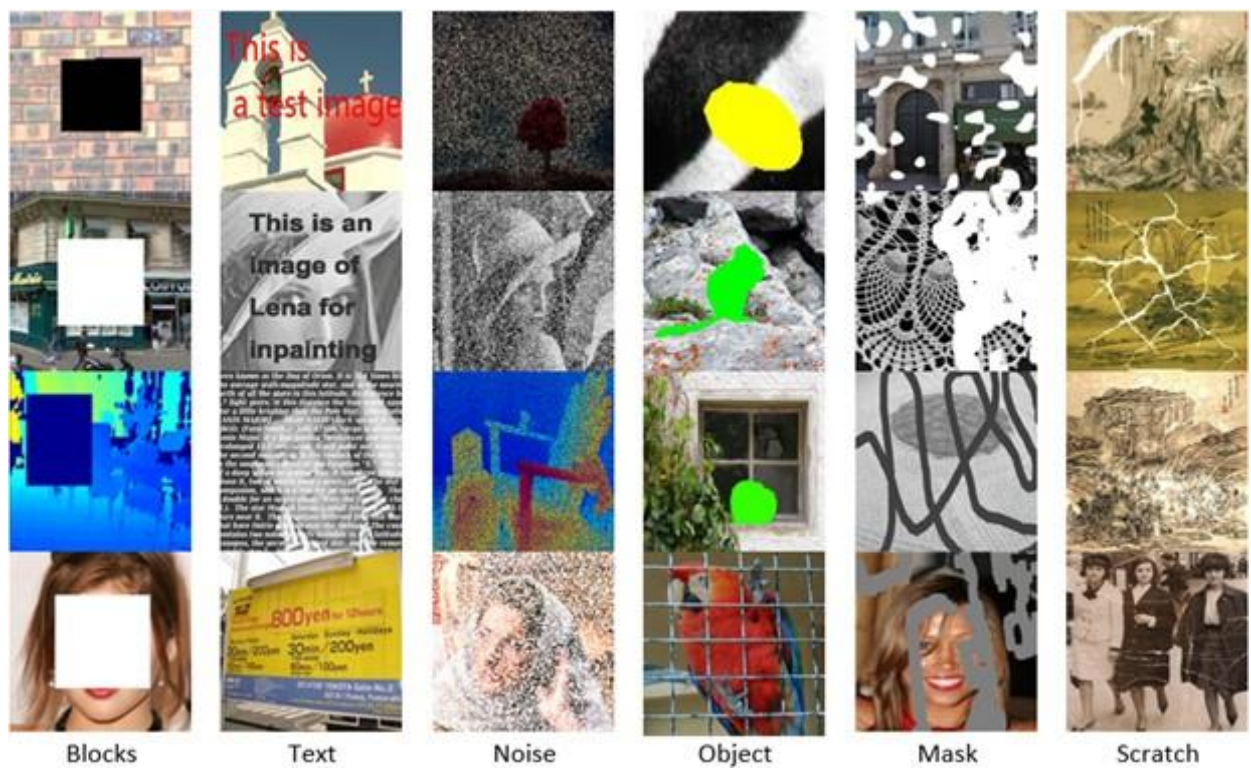

Figure 1: Types of distortion.

Shared images in the social networks can contain many objects added to these images including signature, rectangles or emoticons. The addition of these objects can change the semantic of the images. Removing these objects from the images is a widely recognized problem and a current track in computer vision research. Also, object removal is considered a solution for forgery of images. The object removal techniques that exist in literature can be divided into two categories: image inpainting and copy-move methods. The copy–move based methods perform the undesired object removal by extracting a part from another image or another region from the same image, then pasting it to the object region that we are trying to remove. This technique is widely used for object removal due its simplicity, but it is not suitable for some cases like face images or complicated scenes. Image inpainting was applied on old images in order to remove scratches and enhance damaged images. Now, it is used for removing artifact objects that can be added to the images by filling the target region with estimated values. Image inpainting is also used to remove any type of distortion including text, blocks, noise, scratch, lines or many types of masks [3-5]. Figure 1 represents the different existing types of distortion. By using recently developed algorithms, image inpainting can restore coherently both texture and structure components of the image. The obtained results demonstrate that these methods can remove undesirable objects from the images without leaving traces like artifacts ghosts. Until now, a few methods are proposed for blind image inpainting regarding the massive number of published works with different techniques like sequential-based, CNN-based or GAN-based.

Removing objects from images using image inpainting can reach improved performance in the future, but when the image editors hide their traces using sophisticated techniques, the detection of forgery and the inpainting of image can become difficult. For that reason, almost all detection approaches attempt to handle this by detecting the abnormalities of similarity between blocks of the image that can be affected during the postprocessing operation. For that, For that, this work is a summarization of different methods for image inpainting using different techniques including sequential-based, CNN-based or GAN-based methods.

The remainder of the paper is organized as follows. The literature overview including sequential-based, CNN-based and GAN-based methods are presented in section 2.the used datasets are presented in section 3. Evaluations and metrics used are discussed in section 4. The conclusion is provided in section 5.

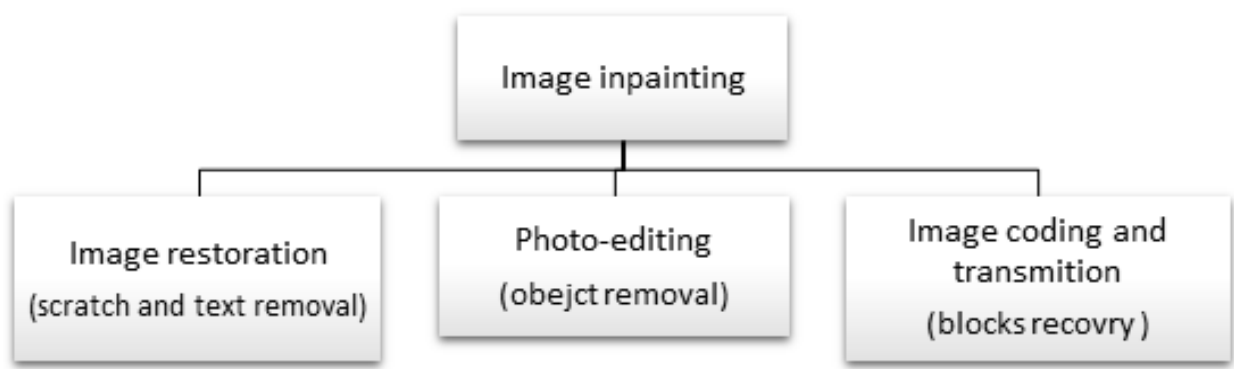

Figure 2: Image inpainting applications and the purposes of each category.

## 2. Related works

Image inpainting is the process of completing or recovering the missing region in the image or removing some object added to it. So, the operation of inpainting depends on the type or domain of applications. For example, in image restoration, we talk about removing the scratch or text that can be found in the images, whereas, in photo-editing application, we are interested in object removal [1], in image coding and transmission applications, the operation related to images inpainting is recovering the missing blocks. Finally, for virtual paintings' restoration, the related operation is crack removal [2]. Figure 2 is a representation of the kind of application related to the corresponding image inpainting operation. To handle this, many methods has been proposed including sequential algorithms or deep learning techniques. By the following a set of existing methods will be presented.

*September 13, 2019*

*2.1. Sequential-based approaches*

The approaches related to images inpainting can classified into two categories: patch-based and diffusion-based methods.

Patch-based methods are based on techniques to fill in the missing region patch-by-patch by searching for well-matching replacement patches (i.e., candidate patches) in the undamaged part of the image and copying them to corresponding locations. Many methods have been proposed for image inpainting using patch-based method. Ružić and Pižurica [3] proposed a patch-based method consisting of searching the well-matched patch in the texture component using Markov Random Field (MRF). Jin and Ye [4] proposed a patch-based approach based on annihilation property filter and low rank structured matrix. In order to remove object from an image, Kawai et al. [5] proposed an approach based on selecting the target object and limiting the search around the target by the background around. Using Two-Stage Low Rank Approximation (TSLRA) [6] and gradient-based low rank approximation [7], authors proposed patch-based methods for recovering the corrupted block in the image. On RGB-D images full of noise and text, Xue et al. [8] proposed a depth image inpainting method based on Low Gradient Regularization. Liu et al. [9] use the statistical regularization and similarity between regions to extract dominant linear structures of target regions followed by repairing of the missing regions using Markov random field model (MRF). Ding et al. [10] proposed a patch-based method for image inpainting using Non-local Texture Matching and Nonlinear Filtering (Alpha-trimmed mean filter). Duan et al. [11] proposed an image inpainting approach based on non-local Mumford–Shah model (NL-MS). Fan and Zhang [12] proposed another image inpainting method based on measuring the similarity between patches using Sum of Squared Differences (SSD). In order to remove blocks from an image, Jiang [13] proposed a method for image compression. Using Singular value decomposition and an approximation matrix, Alilou and Yaghmaee [14] proposed an approach to reconstruct the missing regions. Other notable research includes using texture analysis on Thangka images to recover missing block in an image [15], and using the structure information of images [16,17]. In the same context, Zeng et al. [18] proposed the use of Saliency Map and Gray entropy. Zhang et al. [19] proposed an image inpainting method using joint probability density matrix (JPDM) for object removal from images.

Diffusion-based methods fill in the missing region (the "hole") by smoothly propagating image content from the boundary to the interior of the missing region. For that, Li et al. [20] proposed a diffusion-based method for image inpainting by localizing the diffusion of inpainted regions following by a construction of a feature set based on the intra-channel and inter-channel local variances of the changes to identify the inpainted regions. Another diffusion-based method of image inpainting proposed by the same authors in a later research [21] involves exploiting diffusion coefficients which are computed using the distance and direction between the damaged pixel and its neighborhood pixel. Sridevi et al. [22] proposed another diffusion-based image inpainting method based on Fractional-order derivative and Fourier transform. Table 1 depicts a summary of patch-based and diffusion-based sequential methods for image inpainting.

Jin et al. [23] proposed an approach called sparsity-based image inpainting detection based on canonical correlation analysis (CCA). Mo and Zhou [24] present a research based on dictionary learning using sparse representation. These methods are robust for simple images, but when the image is complex like contains a lot of texture and object or the object cover a large region in the images, searching for similar patch can be difficult.

Table 1: Sequential-based method for image inpainting.

| Category | Method | Feature | Image |
|---|---|---|---|
| Patch-based | Muddala et al. 2016 [1] | layered depth image (LDI) | RGB |
| | ISOGAWA et. 2018 [2] | Super-pixel | RGB |
| | Ružić et al. 2014 [3], Liu et al. 2018 [9] | Markov Random Field (MRF) | RGB |
| | Jin et al. 2015 [4] | Annihilation property filter, low rank structured matrix | RGB |
| | Kawai et al. 2015 [5] | Background geometry estimation | Artificial |
| | Guo et al. 2018 [6] | Two-Stage Low Rank Approximation (TSLRA) | RGB |
| | Xue et al. 2017 [7] | Low Gradient Regularization | Depth |
| | Ding et al. 2019 [11] | Nonlocal Texture Matching, Nonlinear Filtering ($\alpha$-trimmed mean filter) | RGB |
| | Duan et al. 2015 [12] | non-local Mumford–Shah model (NL-MS) | RGB |
| | Fan et al. 2018 [13] | Sum of Squared Differences (SSD) | Gray scale |
| | Jiang et al. 2016 [14] | Canny filter, Segmentation | RGB |
| | Alilou et al. 2017 [15] | Singular value decomposition and an approximation matrix | RGB |
| | Lu et al. 2018 [17] | Gradient-based low rank approximation | Gray scale |
| | Wang et al. 2017 [20], Yao et al. 2018 [22] | Structure and texture analysis | Thangka images |
| | Wei et al. 2016 [21] | Structure-aware | RGB |
| | Zeng et al. 2019 [23] | Saliency Map and Gray entropy | RGB |
| | Zhang et al. 2018 [24] | joint probability density matrix (JPDM). | RGB |
| Diffusion-based | Li et al. 2017 [8] | Intra-channel and Inter-channel local variances | RGB |
| | Li et al. 2016 [16] | Distance and direction between the damaged pixel and its neighborhood pixels. | RGB |
| | Sridevi et al. 2019 [19] | Fractional-order derivative and Fourier transform | Gray scale |

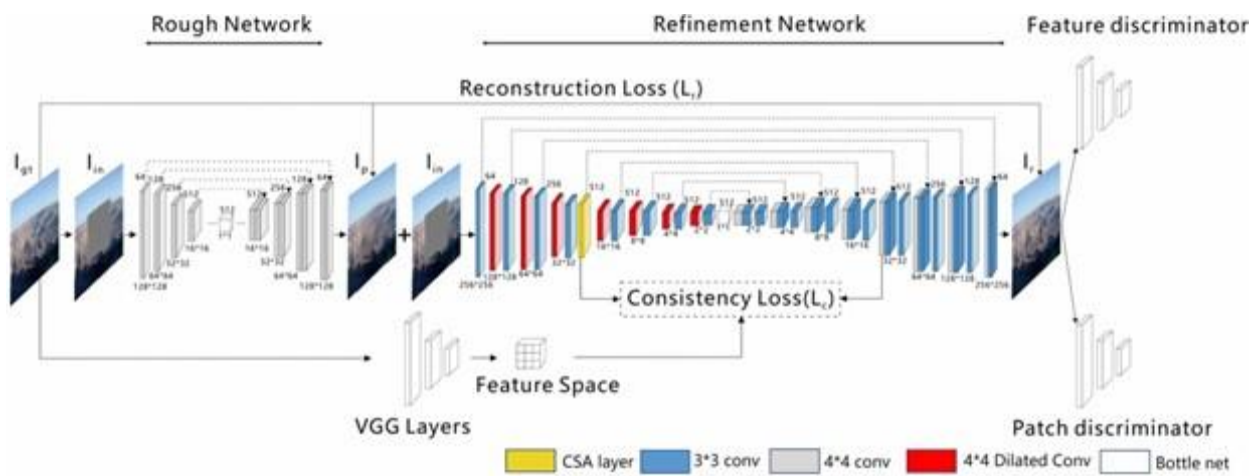

Figure 3: Encoder-decoder networks model in [33].

## 2.2. CNN-based approaches

Recently, the strong potential of deep convolutional networks (CNNs) is being exhibited in all computer vision tasks, especially in image inpainting. CNNs are used specifically in order to improve the expected results in this field using large-scale training data. The sequential-based methods succeed in some part of image inpainting like filling texture details with promising results, but still the problem of capturing the global structure remains [25]. Several methods have been proposed for image inpainting using convolutional neural networks (CNNs) or encoder-decoder network based on CNN. Shift-Net based on U-Net architecture is one of these methods that recover the missing block with good accuracy in terms of structure and fine-detailed texture [25]. In the same context, Weerasekera et al. [26] use Depth Map of the image as input of the CNN architecture, whereas Zhao et al. [27] use the proposed architecture for inpainting X-ray medical images. VORNET [28] is another CNN-based approach for video inpainting for object removal. Most image inpainting methods know the reference of damaged pixels of blocks. Cai et al. [29] proposed a blind image inpainting method named (BICNN). Based on convolutional neural networks (CNNs) using encoder–decoder network structure many works have been proposed for image inpainting. Zhu et al. [30] propose patch-based inpainting method for forensics images. Using the same technique of encoder–decoder network, Sidorov and Hardeberg [31] proposed an architecture for denoising, inpainting and super-resolution for noised, inpainted and low-resolution images, respectively. Zeng et al. [32] built a pyramidal-context architecture called PEN-NET for high-quality image inpainting. Liu et al. [33] proposed a layer to the encoder-decoder network called coherent semantic attention (SCA) layer for image inpainting method. This

*September 13, 2019*

proposed architecture is presented in the figure 3. Further, Pathak et al. [34] proposed using encoder-decoder-based technique for image inpainting. In order to complete the gap between lines drawing in an image, Sasaki et al. [35] use an encoder-decoder-based method. This work can be helpful for scanned data that can miss some parts. For the UAV data that can be affected in terms of resolution or containing some blindspots, Hsu et al. [36] proposed a solution using VGG architecture. Also, for removing some text from the images Nakamura et al. [37] proposed a text erasing method using CNN. In order to enhance the damaged artworks images, Xiang et al. [38] also proposed a CNN-based method. In the same context as [38] and using GRNN neural network, Alilou and Yaghmaee [39] proposed a non-texture image inpainting method. Unlike the previous methods, Liao et al. [40] proposed a method called Artist-Net for image inpainting. The same goal is reached by Cai et al. [41] who proposed a semantic object removal approach using CNN architecture. In order to remove motifs from single images, Hertz et al. [42] proposed a CNN-based approach. Table 2 summarizes CNN-based method with a description of the type of data used for image inpainting.

Table 2: CNN-based method for image inpainting.

| Method | Architecture | Data Type |
|---|---|---|
| Yan et al. 2018 [25], Chang et al. 2019 [28], Zhu et al 2018 [30], Liu et al. 2019 [33], Hsu et al. 2017 [36], Liao et al. 2019 [40], Hertz et al. 2019 [42] | Encoder-decoder network | RGB |
| Weerasekera et al 2018 [26] | CNN | TGB-D |
| Zhao et al. 2018 [27] | CNN | Y-ray |
| Cai et al. 2017 [29] | Blind CNN | Grayscale |
| Sidorov et al. 2019 [31] | 3D CNN | RGB |
| Zeng et al. 2019 [32] | Pyramid-context encoder network | RGB |
| Pathak et al. 2016 [34] | Context-encoder, CNN | RGB |
| Sasaki et al 2017 [35] | CNN | line drawing images |
| Nakamura et al. 2017 [37] | CNN | Image with text |
| Xiang et al. 2017 [39] | CNN, GAN | Damaged old pictures |
| Cai et al. 2018 [41] | CNN | RGB |

## 2.3. GAN-based approaches

GANs are a framework which contains two feed-forward networks, a generator G and a discriminator D, as shown in figure 4. The generator takes random noise as input and generates some fake samples similar to real ones; while the discriminator has to learn to determine whether samples are real or fake. At present, Generative Adversarial Network (GAN) becomes the most used technique in all computer vision applications. GAN-based approaches use a coarse-to-fine network and contextual attention module gives good performance and is proven to be helpful for inpainting [43-47]. Existing image inpainting methods based on GAN are generally a few. Out of these, we find that in [34], Chen and Hu propose GAN-based semantic image inpainting method, named progressive inpainting, where a pyramid strategy from a low-resolution image to higher one is performed for repairing the image. For handwritten images, Li et al. [44] propose a method for inpainting and recognition of occluded characters. The methods use improved GoogLeNet and deep convolutional generative adversarial network (DCGAN). In a research work named PEPSI++ [45], Shin et al. propose a GAN-based method for image inpainting. Wang et al. [46] use Encoder-decoder network and multi-scale GAN for image inpainting. The same combination is used in [47] for image inpainting and image-to-image transformation purposes. On the RBG-D images, Dhamo et al. [48] use CNN and GAN model to generate the background of a scene by removing the object in the foreground image as performed by many methodsod of motion detection using background subtraction [49-51] . In order to complete the missing regions in the image, Vitoria et al. [52] proposed an improved version of the Wasserstein GAN with the incorporation of Discriminator and Generator architecture. In the same context, but on sea surface temperature (SST) images, the Dong et al. [53] proposed a Deep Convolutional Generative adversarial network (DCGAN) for filing the missing parts of the images. Also, Lou et al. [54] exploit a modifier GAN architecture for image inpainting whereas, Salem et al. [55] proposed a semantic image inpainting method using adversarial loss and self-learning encoder-decoder model. A good image restoration method demands to preserve the structural consistency and the texture clarity. For this reason, Liu et al. [56] proposed a GAN-based method for image inpainting on face images. FiNet [57] is another approach found in literature for fashion image inpainting that consists of completing the missing parts in fashion images.

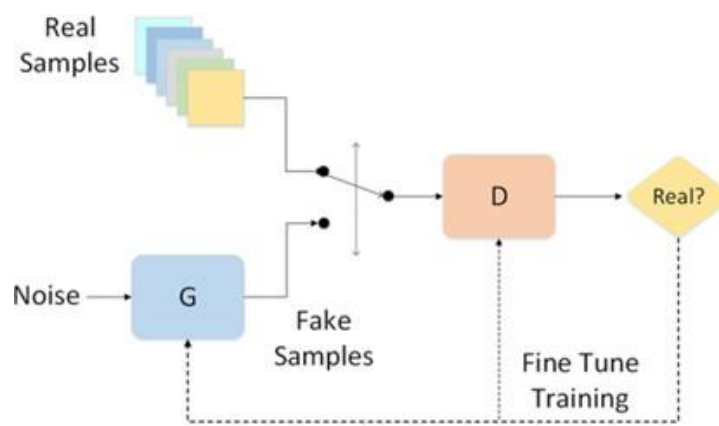

Figure 4: Framework of GANs.

The GAN-based methods give a good addition to the performance of image inpainting algorithms, but the speed of training is lower and needs very good performance machines, and this is due to computational resources requirements including network parameters and convolution operations.

## 3. Image inpainting datasets

Image inpainting methods use many public and large datasets for evaluating their algorithms and comparing the performance. The categories of images determine the effectiveness of each proposed method. From these categories we can find natural images, artificial images, face images, and many other categories. In this work, we attempt to collect the most used datasets for image inpainting including Paris StreetView [58], Places [59], depth image dataset [8], Foreground-aware [60], Berkeley segmentation [61], ImageNet [62] and others. We also try to cite the types of used data such as RGB images, RGB-D images and SST images. Figure 5 represents some frame examples from the cited datasets. Where Table 3 describe various datasets used for image inpainting approaches.

**Paris StreetView** [58] is Collected from Google StreetView that represent a large-scale dataset that contains street images for several cities around the world. The Paris StreetView composed of 15000 images. The image's resolution is for 936 $\times$537 pixels.

**Places** [1] [59] datasets built for human visual cognition and visual understanding purposes. The dataset contains many scene categories such as bedrooms, streets, synagogue, canyon and others. The dataset is composed of 10 million images including 400> image for each scene category. It allows the deep learning methods to train their architecture with a large-scale data

**Depth image** dataset[2] is introduced by [8] for evaluating depth image inpainting methods. The dataset is composed of two types RGB-D images and grayscale depth images. Also, 14 scene categories are included such as Adirondack, Jade plant, Motorcycle, Piano, Playable and others. The masks for damaged images are created including textual makes (text in the images) and random missing masks.

**Foreground-aware** dataset [60] is different from the other's dataset. It contains the masks that can be added to any images for damaging it. It named irregular hole mask dataset for image inputting. Foreground-aware datasets contains 100,000 masks with irregular holes for training, and 10,000 masks for testing. Each mask is a 256$\times$256 gray image with 255 indicating the hole pixels and 0 indicating the valid pixels. The masks can be added to any image for which can be used for creating a large dataset of damaged images.

**Berkeley segmentation database**[3] [61] is composed of 12 000 images segmented manually. The images collected from other dataset contains 30 human subjects. The dataset is a combining of RGB images and Grayscale images.

---

[1]http://places2.csail.mit.edu/download.html  
[2]http://www.cad.zju.edu.cn/home/dengcai/Data/depthinpaint/DepthInpaintData.html  
[3]https://www2.eecs.berkeley.edu/Research/Projects/CS/vision/bsds/

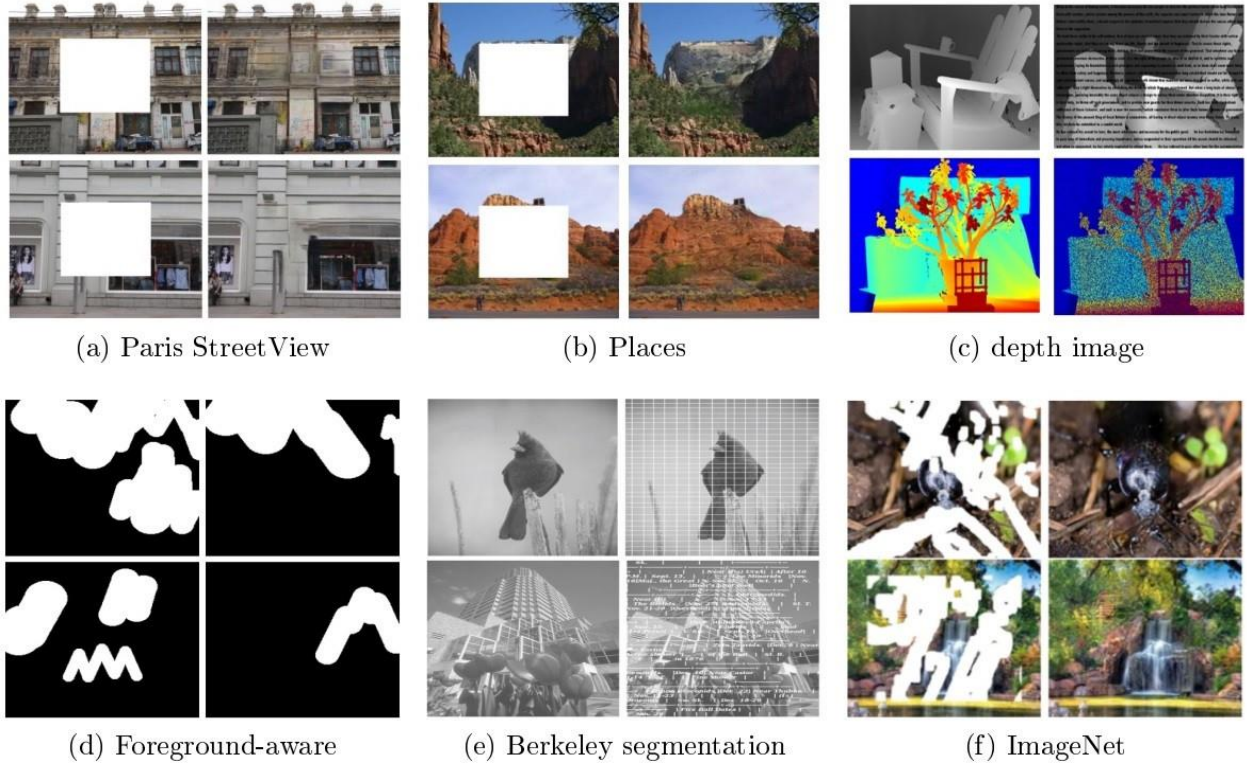

Figure 5: Examples from Image inpainting datasets

**ImageNet**[4] [62] is a large-scale dataset with thousands of images of each subnet. Each subnet is presented of 1000 images. The current version of the dataset contains more than 14,197,122 images where the 1,034,908 annotated with bounding box human body is annotated.

**USC-SIPI image database**[5] contains several volumes representing the many types of images. The resolution in each volume can vary between 256×256, 512×512 and 1024×1024 pixels. Generally, the datasets contain 300 images representing four volumes including texture, aerials, Miscellaneous and sequences.

**CelebFaces Attributes Dataset (CelebA)**[6] [63] is a recognized and public datasets for face recognition.it contains more that 200K celebrity images representing 10 000 identities with a large pose variations.

**Indian Pines**[7][64] consist of images representing images of three scenes including agriculture, forest and natural perennial vegetation with resolution of 145×145 pixels.

**Microsoft COCO val2014 dataset**[8][65] is a new image recognition, segmentation,

---

[4]http://image-net.org/
[5]http://sipi.usc.edu/database/
[6]http://mmlab.ie.cuhk.edu.hk/projects/CelebA.html
[7]http://www.ehu.eus/ccwintco/index.php/Hyperspectral_Remote_Sensing_Scenes
[8]http://cocodataset.org/#home

Table 3: Datasets description

| Dataset | Year | Content category | Image Type | Size and resolution | Papers |
|---|---|---|---|---|---|
| Paris street view | 2015 | Street | RGB | 936 × 537 | [25][33][45][46][56] |
| Places | 2017 | Miscellaneous | RGB | 10 Mil, 256×256 | [25][30][32][45] |
| Foreground-aware | 2019 | Masks | Binary, grayscale | 100K, 256×256 | [7] |
| Berkeley segmentation | 2001 | Miscellaneous | RGB | 12K | [4][10] [[29] |
| ImageNet | 2015 | Miscellaneous | RGB | 100K , variant | [29][40][41][45] |
| CelebA | 2018 | Face images | RGB | 200K, variant | [43][45][52][56] |
| Indian Pines | 2015 | Agriculture and forest scenes | RGB, gray scale | 145x145 | [31] |
| Microsoft COCO val2014 | 2014 | Miscellaneous | RGB | 2.5 Mil, 256×256 | [42] |
| ICDAR 2013 | 2013 | RGB images with text | RGB | 27 GB, variant | [64] |
| SceneNet | 2016 | Miscellaneous | RGB, RGB-D | 5Mil, variant | [48] |
| Stanford Cars | 2013 | Cars | RGB | 16K | [43] |
| Cityscapes | 2016 | Miscellaneous | RGB | 48GB, Variant | [47] |
| Middlebury Stereo | 2006 2014 | Miscellaneous | RGB, RGB-D, Grayscale depth images | 1396 x 1110 (2006) 2964 x 1988 (2014) | [7][8] |

and captioning dataset. Microsoft COCO has several features with a total of 2.5 million labeled instances in 328k images.

Benchmark **dataset ICDAR 2013**[9][66] is a handwritten datasets with two languages including Arabic and English. The total number of writer is 475 where the images have been scanned. The datasets contain 27 GB of data.

**SceneNet dataset**[10][67] is a dataset for scene understanding tasks including semantic segmentation, object detection and 3D reconstruction. It contains RGB image and the corresponding RGB-D images which are in total 5 million images.

---

[9]http://www.tamirhassan.com/html/competition.html
[10]https://robotvault.bitbucket.io/scenenet-rgbd.html

**Stanford Cars dataset**[11][68] is a set of cars represent 196 categories of cars with different size. The datasets contain 16 200 images in total.

**Cityscapes dataset** [12][69] is large scale dataset of stereo videos of street scene of 50 cities. The images contain about 30 classes of objects. Also, about 20 000 annotated frames with coarse annotations.

**Middlebury Stereo**[13] datasets contain many versions we present the two new ones [70] and [71]. Middlebury 2006 [70] is a depth grayscale dataset that contains images captured from 7 view with different illuminations and exposures. The images resolution is defined by three categories full-size with 1240× 1110 pixels, half size with 690× 555 pixels and the third resolution with 413 ×370. Middlebury 2014 [71] is an RGB-D datasets unlike the other version.

## 4. Evaluation and discussion

Due to the unavailability of a large dataset of damaged painting images and due also to the novelty of the image inpainting topic, researchers find it difficult to obtain datasets for training their methods [72]. For that, most researchers use the existing datasets like USC-SIPI, Paris StreetView, Places, ImageNet and others, and damage a set of images from these datasets for training their models and algorithms. The existing methods in literature generate their own image inpainting datasets by adding some artificial distortion including noise [8], text [12], scratch [18], objects (shapes) [58], masks [60,62].

The method of evaluation of the obtained results for image inpainting algorithms differs regarding the technique used. For the sequential-based approaches metrics used for evaluation are unlike CNN-based methods. In order to evaluate efficiency of the proposed methods, researchers use some evaluation measures including mean squared error (MSE), Peak Signal to Noise Ratio (PSNR) and Structural Similarity Index (SSIM) [73]. For example, Zeng et al. [18] use these evaluation parameters to demonstrate the obtained results of repairing of scratch and text in the images. Whereas Mo et al. [24] use the same metrics for evaluating the experiments for text and noise. In the same context, Duan et al. [11] use the same metrics for testing the proposed method of removing the added objects to the images. In addition to the metrics used for evaluation, the category of image in the used dataset for evaluation can be different from one method to another. For example, some methods use RGB images while others evaluate their methods in RGB-D images or historical images. For that reason, we attempt to summarize the obtained results regarding each category of images used and the type of damaging in the images. In addition, this summarization is for sequential-based methods, which use the common evaluation metrics as shown in Table 4. Form the table we can observe that most of these methods are evaluated on grayscale images like in [4, 6, 7, 12, 13, 22, 24] with some type of distortion including text, for example, Gaussian noise (named Random in the papers) or some type of objects. We can also find some methods to analyze the three types like [7], whereas others process just for two types (text and noise) like in [6, 22, 24]. Also from the table, we can detect that the proposed methods use some recognized images in computer vision like Lena and Barbara images, which are used for testing the effectiveness of several methods. Methods that propose their approaches for image inpainting on RGB images use the same distortion categories including text, noise, and objects [1, 3, 10, 11, 18]. In addition, some researchers propose methods for scratch analysis, which is a process of restoring the old images or images with damaged by some lines like in [3] and [17]. As shown in the tables, the state-of-the-art methods use different images from internet or some datasets and this is because of the lack of datasets for image inpainting. The third type of images used in literature for image inpainting is RGB-D whereby some researchers use an image inpainting approach on RGB-D image.

---

[11]https://ai.stanford.edu/~jkrause/cars/car_dataset.html
[12]https://www.cityscapes-dataset.com/
[13]http://vision.middlebury.edu/stereo/data/

Table 4: Summarization of sequential-based methods evaluations

| Image type | Type of distortion | Method | PSNR | Images source |
|---|---|---|---|---|
| RGB | text | Zeng et al. 2019 [23] | 42.74 | Internets images |
| | | Ružić et al. 2015 [3] | 28.20 | Internet images |
| | Random objects | Duan et al. 2015 [12] | 27.40 | Internet images |
| | | Muddala et al. 2016 [1] | 28.52 | Internet images |
| | | Ding et al. 2019 [11] | 29.19 | Berkeley segm |
| | | Duan et al. 2015 [12] | 31.94 | Self-collected |
| | | Zeng et al. 2019 [23] | 36.98 | Internet images |
| | Scratch | Ružić et al. 2015 [3] | 23.89 | Internet images |
| | | Yao et al. 2018 [22] | 24.31 | Thangka |
| RGB-D | Text | Xeu et al. 2017 [7] | 27.65 | Middlebury stereo |
| | Random | Xeu et al. 2017 [7] | 27.18 | Middlebury stereo |
| Grayscale | Text | Fan et al. 2018 [13] | 32.83 | Internet images |
| | | Lu et al. 2018 [17] | 40.35 | Barbara |
| | | Mo et al. 2018 [18] | 35.90 | Barbara |
| | | Geo et al. 2018 [6] | 23.79 | Lena |
| | | Sridevi et al. 2019 [19] | 40.63 | Lena |
| | Random | Jin et al. 2015 [4] | 38.04 (50%) | Barbara |
| | | Jin et al. 2015 [4] | 38.09 (50%) | Lena |
| | | Geo et al. 2018 [6] | 29.39 (40%) | Lena |
| | | Lu et al. 2018 [17] | 30.31, (25%) | Barbara |
| | | Mo et al. 2018 [18] | 39.65 (25%) | Barbara |
| | | Sridevi et al. 2019 [19] | 37.51 (40%) | Lena(512x512) |
| | objects | Geo et al. 2018 [6] | 29.18 | Lena |
| | | Fan et al. 2018 [13] | 22.63 | Lena |
| | | Jiang et al. 2016 [14] | 31.8 | Lena |
| | | Lu et al. 2018 [17] | 39.05 | Lena |

Table 5: Performance of CNN-based methods

| Method | Dataset | Distortion type | PSNR | SSIM |
|---|---|---|---|---|
| [25] | Paris StreetView Places | Random region | 26.51 | 0.9 |
| [29] | Berkley segmentation | Horiz/verti lines | 38.22 | 0.9885 |
|  |  | Text | 36.21 | 0.9741 |
|  |  | Text (Lena) | 36.67 | 0.9714 |
| [31] | Indian Pines | Vertical lines | 37.54 | 0.979 |
| [32] | Places | Region blocks | - | 0.7809 |
| [33] | Paris StreetView | Random region | 32.67 | 0.972 |
|  | CelebA | Random region | 34.69 | 0.989 |
| [34] | Paris Street | Random region | 17.59 | - |
| [36] | ImageNet | Random region | 25.43 | 0.69 |
| [38] | Internet images | Scratch (old images) | 27.42 | 0.9481 |
| [39] | Internet images | Lena (scratch) | 39.64 | - |
|  |  | Lena (text) | 36.38 | - |
|  |  | Lena (Noise 50%) | 29.59 | - |
|  |  | Lena (Noise 70%) | 24.67 | - |
| [40] | Paris StreetView | Scratch | 19.94 | 0.59 |
|  | ImageNet | Region block | 17.67 | 0.52 |
| [41] | Paris StreetView ImageNet | Random region | 6.69 | 0.19 |
| [42] | Microsoft COCO-val2014 | Text and symbols | 37.63 | 0.9889 |
| [27] | Lung CT Test | Blocks | 33.86 | 0.9781 |
|  | ZUBAL CT |  | 36.33 | 0.9838 |

With deep learning techniques, any task in computer vision can be performed with an automatic learning using different unsupervised features, unlike the sequential-based method. The learning is made using convolutional neural networks (CNNs) that makes several computer vision tasks more improved in terms of robustness and most simple in terms of features suitable for each task. For image inpainting methods that use CNN as described in section above, the effectiveness of each approach is related to the size and type of the data used and the architecture implemented. The evaluation of these methods is the same as for sequential-based methods. PSNR (the distance at the pixel level) and SSIM (similarity between two images) are used for evaluating the robustness of repairing the damaged images under different categories of distortion including scratch, text, noise or random region (Blocks) added to the image. Related to this, Table 5 represents CNN-based methods for image inpainting and their performance evaluation and describes the datasets used, the type of distortion, evaluation metrics and the resolution of the images used in training. It is obvious that the performance of such methods is related to the type of distortion. For example, the images damaged by blocks are less accurate in term of PSNR values. The algorithms [29, 31, 39, 42] can handle the added visual motifs like text or lines with a good performance. In addition, the performance is influenced by the percentage of added noise to the images. For the new dataset used for image inpainting, including Paris StreetView, Places or ImageNet that contains a large scale of data which is also different types of images, the algorithms accuracy can be less than the others approaches using another dataset [25, 34, 40, 41]. This change in accuracy is related to the diversity of the images in these datasets.

Table 6: GAN-based performance results.

| Method | Datasets | Type distortion | PSNR | SSIM | Resolution |
|--------|----------|-----------------|------|------|------------|
| [43] | CelebA | Center block | 21.45 | 0.851 | 64×64 |
|  | Stanford Cars dataset |  | 15.02 | 0.725 |  |
| [45] | CelebA | Blocks | 25.6 | 0.901 | 256 × 256 |
|  |  | Free-form mask | 28.6 | 0.929 |  |
|  | Places | Blocks | 21.5 | 0.839 |  |
|  |  | Free-form mask | 25.2 | 0.889 |  |
| [47] | Cityscapes dataset | Region blocks | 16.06 | 0.4820 | 256 × 256 |
| [49] | CelebA | Random region | 23.06 | 0.9341 | 64×64 |
|  | SVHN dataset |  | 26.23 | 0.8969 |  |
| [48] | SceneNet dataset | Masks (RGB-D) | 22.35 | 0.891 | - |

Each method either makes a visual or qualitative evaluation; or an evaluation using metrics or quantitative evaluation. The quantitative evaluation, using PSNR and SSIM metrics, is performed also for image inpainting with GAN-based methods. In some cases, these metrics do not mean that qualitative results are better. This is related to the ground truth that should be unique [43]. Also, some methods for image inpainting are better for a certain category of images as well as the type of distortion used. Table 6 shows a number of GAN-based methods for images processing with a description of the datasets used and the evaluation metrics used for each method. In [47] the evaluation is made using many metrics depending on the position of the damaged region (block) including center, left right, up, and down. But here we choose to present just PSNR and SSSIM of the image inpainting results on the images where the block is located in the center. In [45], two datasets are used with two types of distortion including blocks and free-form masks, which are categories of scratch painted with bold lines. For this example, we can see that the inpainting of scratch is more accurate that repairing the blocks . This becomes obvious from the fact that the blocks can take a region of the images where the scratch can take distributed small regions in the image.

As mentioned above, the unavailability of the datasets for image inpainting make the comparison between these methods difficult. Also, each author uses different masks and type of distortion.

## 5. Conclusions

Image inpainting is an important task for computer vision applications, due to large modified data using images editing tools. From these applications, we can find wireless image coding and transmission, image quality enhancement, image restoration and others. In this paper, a brief image inpainting review is performed. Different categories of approaches have been presented including sequential-based (classical approach without learning), CNN-based approach and GAN-based approaches. We also attempt to collect the approaches that handle different types of distortion in the images such as text, objects added, scratch, and noise as well as several categories of data like RGB, RGB-D, historical images. A good alternative to these conventional features is the learned ones, e.g. deep learning, which has more generalization ability in more complicated scenarios. To be effective, these models need to be trained on a large amount of data. For that, we collect the most used datasets used for training these models. In order to summarize the different analyzed cases and their performance, we present a description using tables for each category of methods by presenting their evaluation performing the types of data, the datasets and the metrics used for each approach.

As a conclusion, there is no method that can inpaint all the types of distortion in images, but using learning techniques there are some promising results for each category of analyzed cases.

**Acknowledgments**

This publication was made by NPRP grant \# NPRP8-140-2-065 from the Qatar National Research Fund (a member of the Qatar Foundation). The statements made herein are solely the responsibility of the authors.

**References**

1. S. M. Muddala, R. Olsson, M. Sjöström, Spatio-temporal consistent depth-image-based rendering using layered depth image and inpainting, EURASIP Journal on Image and Video Processing 2016 (1) (2016)9.
2. M. Isogawa, D. Mikami, D. Iwai, H. Kimata, K. Sato, Mask optimization for image inpainting, IEEE Access 6 (2018) 69728-69741.
3. T. Ruºi¢, A. Piºurica, Context-aware patch-based image inpainting using markov random field modeling, IEEE Transactions on Image Processing 24 (1) (2014) 444-456.
4. K. H. Jin, J. C. Ye, Annihilating filter-based low-rank hankel matrix approach for image inpainting,IEEE Transactions on Image Processing 24 (11) (2015) 3498-3511.
5. N. Kawai, T. Sato, N. Yokoya, Diminished reality based on image inpainting considering background geometry, IEEE transactions on visualization and computer graphics 22 (3) (2015) 1236-1247.

6. Q. Guo, S. Gao, X. Zhang, Y. Yin, C. Zhang, Patch-based image inpainting via two-stage low rank approximation, IEEE transactions on visualization and computer graphics 24 (6) (2017) 2023-2036.
7. H. Lu, Q. Liu, M. Zhang, Y. Wang, X. Deng, Gradient-based low rank method and its application in image inpainting, Multimedia Tools and Applications 77 (5) (2018) 5969-5993.
8. H. Xue, S. Zhang, D. Cai, Depth image inpainting: Improving low rank matrix completion with low gradient regularization, IEEE Transactions on Image Processing 26 (9) (2017) 4311-4320.
9. J. Liu, S. Yang, Y. Fang, Z. Guo, Structure-guided image inpainting using homography transformation,IEEE Transactions on Multimedia 20 (12) (2018) 3252-3265.
10. D. Ding, S. Ram, J. J. Rodríguez, Image inpainting using nonlocal texture matching and nonlinear filtering, IEEE Transactions on Image Processing 28 (4) (2018) 1705-1719.
11. J. Duan, Z. Pan, B. Zhang, W. Liu, X.-C. Tai, Fast algorithm for color texture image inpainting using the non-local ctv model, Journal of Global Optimization 62 (4) (2015) 853-876..
12. Q. Fan, L. Zhang, A novel patch matching algorithm for exemplar-based image inpainting, MultimediaTools and Applications 77 (9) (2018) 10807-10821.
13. W. Jiang, Rate-distortion optimized image compression based on image inpainting, Multimedia Tools and Applications 75 (2) (2016) 919-933..
14. V. K. Alilou, F. Yaghmaee, Exemplar-based image inpainting using svd-based approximation matrix and multi-scale analysis, Multimedia Tools and Applications 76 (5) (2017) 7213-7234.
15. W. Wang, Y. Jia, Damaged region filling and evaluation by symmetrical exemplar-based image inpainting for thangka, EURASIP Journal on Image and Video Processing 2017 (1) (2017) 38..
16. Y. Wei, S. Liu, Domain-based structure-aware image inpainting, Signal, Image and Video Processing10 (5) (2016) 911-919.
17. F. Yao, Damaged region filling by improved criminisi image inpainting algorithm for thangka, ClusterComputing (2018) 1-9..
18. J. Zeng, X. Fu, L. Leng, C. Wang, Image inpainting algorithm based on saliency map and gray entropy,Arabian Journal for Science and Engineering 44 (4) (2019) 3549-3558.
19. D. Zhang, Z. Liang, G. Yang, Q. Li, L. Li, X. Sun, A robust forgery detection algorithm for object removal by exemplar-based image inpainting, Multimedia Tools and Applications 77 (10) (2018) 11823-11842..
20. H. Li, W. Luo, J. Huang, Localization of diffusion-based inpainting in digital images, IEEE Transactionson Information Forensics and Security 12 (12) (2017) 3050-3064.
21. K. Li, Y. Wei, Z. Yang, W. Wei, Image inpainting algorithm based on tv model and evolutionary algorithm, Soft Computing 20 (3) (2016) 885-893..
22. G. Sridevi, S. S. Kumar, Image inpainting based on fractional-order nonlinear diffusion for image reconstruction, Circuits, Systems, and Signal Processing (2019) 1-16.

23. X. Jin, Y. Su, L. Zou, Y. Wang, P. Jing, Z. J. Wang, Sparsity-based image inpainting detection viacanonical correlation analysis with low-rank constraints, IEEE Access 6 (2018) 49967-49978..
24. J. Mo, Y. Zhou, The research of image inpainting algorithm using self-adaptive group structure and sparse representation, Cluster Computing (2018) 1-9.
25. Z. Yan, X. Li, M. Li, W. Zuo, S. Shan, Shift-net: Image inpainting via deep feature rearrangement, in:Proceedings of the European Conference on Computer Vision (ECCV), 2018, pp. 1-17..
26. C. S. Weerasekera, T. Dharmasiri, R. Garg, T. Drummond, I. Reid, Just-in-time reconstruction: Inpainting sparse maps using single view depth predictors as priors, in: 2018 IEEE International Conference on Robotics and Automation (ICRA), IEEE, 2018, pp. 1-9.
27. J. Zhao, Z. Chen, L. Zhang, X. Jin, Unsupervised learnable sinogram inpainting network (sin) for limited angle ct reconstruction, arXiv preprint arXiv:1811.03911..
28. Y.-L. Chang, Z. Yu Liu, W. Hsu, Vornet: Spatio-temporally consistent video inpainting for object removal, in: Proceedings of the IEEE Conference on Computer Vision and Pattern Recognition Workshops, 2019.
29. N. Cai, Z. Su, Z. Lin, H. Wang, Z. Yang, B. W.-K. Ling, Blind inpainting using the fully convolutional neural network, The Visual Computer 33 (2) (2017) 249-261..
30. X. Zhu, Y. Qian, X. Zhao, B. Sun, Y. Sun, A deep learning approach to patch-based image inpainting forensics, Signal Processing: Image Communication 67 (2018) 90-99.
31. O. Sidorov, J. Y. Hardeberg, Deep hyperspectral prior: Denoising, inpainting, super-resolution, arXiv preprint arXiv:1902.00301..
32. Y. Zeng, J. Fu, H. Chao, B. Guo, Learning pyramid-context encoder network for high-quality image inpainting, in: Proceedings of the IEEE Conference on Computer Vision and Pattern Recognition,2019, pp. 1486-1494.
33. H. Liu, B. Jiang, Y. Xiao, C. Yang, Coherent semantic attention for image inpainting, arXiv preprint arXiv:1905.12384..
34. D. Pathak, P. Krahenbuhl, J. Donahue, T. Darrell, A. A. Efros, Context encoders: Feature learning by inpainting, in: Proceedings of the IEEE conference on computer vision and pattern recognition, 2016,pp. 2536-2544.
35. K. Sasaki, S. Iizuka, E. Simo-Serra, H. Ishikawa, Joint gap detection and inpainting of line drawings,in: Proceedings of the IEEE Conference on Computer Vision and Pattern Recognition, 2017, pp.5725-5733.
36. C. Hsu, F. Chen, G. Wang, High-resolution image inpainting through multiple deep networks, in: 2017International Conference on Vision, Image and Signal Processing (ICVISP), IEEE, 2017, pp. 76-81.
37. T. Nakamura, A. Zhu, K. Yanai, S. Uchida, Scene text eraser, in: 2017 14th IAPR InternationalConference on Document Analysis and Recognition (ICDAR), Vol. 1, IEEE, 2017, pp. 832-837.
38. P. Xiang, L. Wang, J. Cheng, B. Zhang, J. Wu, A deep network architecture for image inpainting, in:2017 3rd IEEE International Conference on Computer and Communications (ICCC), IEEE, 2017, pp.1851-1856.

39. V. K. Alilou, F. Yaghmaee, Application of grnn neural network in non-texture image inpainting and restoration, Pattern Recognition Letters 62 (2015) 24-31.
40. L. Liao, R. Hu, J. Xiao, Z. Wang, Artist-net: Decorating the inferred content with unified style for image inpainting, IEEE Access 7 (2019) 36921-36933.
41. X. Cai, B. Song, Semantic object removal with convolutional neural network feature-based inpaintin gapproach, Multimedia Systems 24 (5) (2018) 597-609.
42. A. Hertz, S. Fogel, R. Hanocka, R. Giryes, D. Cohen-Or, Blind visual motif removal from a singleimage, arXiv preprint arXiv:1904.02756.
43. Y. Chen, H. Hu, An improved method for semantic image inpainting with gans: Progressive inpainting, Neural Processing Letters 49 (3) (2019) 1355-1367..
44. J. Li, G. Song, M. Zhang, Occluded offine handwritten chinese character recognition using deep convolutional generative adversarial network and improved googlenet, Neural Computing and Applications 1-15.
45. Y.-G. Shin, M.-C. Sagong, Y.-J. Yeo, S.-W. Kim, S.-J. Ko, Pepsi++: Fast and lightweight network for image inpainting, arXiv preprint arXiv:1905.09010..
46. H. Wang, L. Jiao, H. Wu, R. Bie, New inpainting algorithm based on simplified context encoders and multi-scale adversarial network, Procedia computer science 147 (2019) 254-263.
47. C. Wang, C. Xu, C. Wang, D. Tao, Perceptual adversarial networks for image-to-image transformation,IEEE Transactions on Image Processing 27 (8) (2018) 4066-4079..
48. H. Dhamo, K. Tateno, I. Laina, N. Navab, F. Tombari, Peeking behind objects: Layered depth prediction from a single image, Pattern Recognition Letters 125 (2019) 333-340.
49. O. ELHarrouss, D. Moujahid, S. E. Elkaitouni, H. Tairi, Moving objects detection based on thresholding operations for video surveillance systems, in: 2015 IEEE/ACS 12th International Conference of Computer Systems and Applications (AICCSA), IEEE, 2015, pp. 1-5..
50. O. Elharrouss, A. Abbad, D. Moujahid, J. Riffi, H. Tairi, A block-based background model for moving object detection, ELCVIA: electronic letters on computer vision and image analysis 15 (3) (2016)0017-31.
51. O. Elharrouss, D. Moujahid, S. Elkah, H. Tairi, Moving object detection using a background modeling based on entropy theory and quad-tree decomposition, Journal of Electronic Imaging 25 (6) (2016)061615..
52. P. Vitoria, J. Sintes, C. Ballester, Semantic image inpainting through improved wasserstein generative adversarial networks, arXiv preprint arXiv:1812.01071.
53. J. Dong, R. Yin, X. Sun, Q. Li, Y. Yang, X. Qin, Inpainting of remote sensing sst images with deep convolutional generative adversarial network, IEEE Geoscience and Remote Sensing Letters 16 (2)(2018) 173-177..
54. S. Lou, Q. Fan, F. Chen, C. Wang, J. Li, Preliminary investigation on single remote sensing image inpainting through a modified gan, in: 2018 10th IAPR Workshop on Pattern Recognition in RemoteSensing (PRRS), IEEE, 2018, pp. 1-6.

55. N. M. Salem, H. M. Mahdi, H. Abbas, Semantic image inpainting vsing self-learning encoder-decoder and adversarial loss, in: 2018 13th International Conference on Computer Engineering and Systems(ICCES), IEEE, 2018, pp. 103-108..
56. H. Liu, G. Lu, X. Bi, J. Yan, W. Wang, Image inpainting based on generative adversarial networks, in:2018 14th International Conference on Natural Computation, Fuzzy Systems and Knowledge Discovery(ICNC-FSKD), IEEE, 2018, pp. 373-378.
57. X. Han, Z. Wu, W. Huang, M. R. Scott, L. S. Davis, Compatible and diverse fashion image inpainting,arXiv preprint arXiv:1902.01096..
58. C. Doersch, S. Singh, A. Gupta, J. Sivic, A. A. Efros, What makes paris look like paris?, Communications of the ACM 58 (12) (2015) 103-110.
59. B. Zhou, A. Lapedriza, A. Khosla, A. Oliva, A. Torralba, Places: A 10 million image database for scene recognition, IEEE transactions on pattern analysis and machine intelligence 40 (6) (2017) 1452-1464..
60. W. Xiong, J. Yu, Z. Lin, J. Yang, X. Lu, C. Barnes, J. Luo, Foreground-aware image inpainting, in:Proceedings of the IEEE Conference on Computer Vision and Pattern Recognition, 2019, pp. 5840-5848.
61. D. Martin, C. Fowlkes, D. Tal, J. Malik, et al., A database of human segmented natural images and its application to evaluating segmentation algorithms and measuring ecological statistics, Iccv Vancouver:,2001..
62. O. Russakovsky, J. Deng, H. Su, J. Krause, S. Satheesh, S. Ma, Z. Huang, A. Karpathy, A. Khosla,M. Bernstein, et al., Imagenet large scale visual recognition challenge, International journal of computer vision 115 (3) (2015) 211-252.
63. Z. Liu, P. Luo, X. Wang, X. Tang, Large-scale celeb faces attributes (celeba) dataset, Retrieved August15 (2018) 2018..
64. M. F. Baumgardner, L. L. Biehl, D. A. Landgrebe, 220 band aviris hyperspectral image data set: June12, 1992 indian pine test site 3, Purdue University Research Repository 10 (2015) R7RX991C.
65. T.-Y. Lin, M. Maire, S. Belongie, J. Hays, P. Perona, D. Ramanan, P. Dollár, C. L. Zitnick, Microsoftcoco: Common objects in context, in: European conference on computer vision, Springer, 2014, pp.740-755..
66. D. Karatzas, F. Shafait, S. Uchida, M. Iwamura, L. G. i Bigorda, S. R. Mestre, J. Mas, D. F. Mota,J. A. Almazan, L. P. De Las Heras, Icdar 2013 robust reading competition, in: 2013 12th InternationalConference on Document Analysis and Recognition, IEEE, 2013, pp. 1484-1493.
67. J. McCormac, A. Handa, S. Leutenegger, A. J. Davison, Scenenet rgb-d: 5m photorealistic images of synthetic indoor trajectories with ground truth, arXiv preprint arXiv:1612.05079.1.
68. J. Krause, M. Stark, J. Deng, L. Fei-Fei, 3d object representations for fine-grained categorization, in:Proceedings of the IEEE International Conference on Computer Vision Workshops, 2013, pp. 554-561.
69. M. Cordts, M. Omran, S. Ramos, T. Rehfeld, M. Enzweiler, R. Benenson, U. Franke, S. Roth, B. Schiele,The cityscapes dataset for semantic urban scene understanding, in: Proceedings of the IEEE conference on computer vision and pattern recognition, 2016, pp. 3213-3223.8.
70. H. Hirschmuller, D. Scharstein, Evaluation of cost functions for stereo matching, in: 2007 IEEE Conference on Computer Vision and Pattern Recognition, IEEE, 2007, pp. 1-8.
71. D. Scharstein, H. Hirschmüller, Y. Kitajima, G. Krathwohl, N. Ne²i¢, X. Wang, P. Westling, High resolution stereo datasets with subpixel-accurate ground truth, in: German conference on pattern recognition, Springer, 2014, pp. 31-42.7.
72. N. H. Jboor, A. Belhi, A. K. Al-Ali, A. Bouras, A. Jaoua, Towards an inpainting framework for visualcultural heritage, in: 2019 IEEE Jordan International Joint Conference on Electrical Engineering andInformation Technology (JEEIT), IEEE, 2019, pp. 602-607.
73. O. Elharrouss, A. Abbad, D. Moujahid, H. Tairi, Moving object detection zone using a block-based background model, IET Computer Vision 12 (1) (2017) 86-94.